\documentclass{article}

\usepackage{microtype}
\usepackage{graphicx}
\usepackage{subfigure}
\usepackage{booktabs} 
\usepackage{amsfonts,amsmath,mathrsfs}       
\usepackage{nicefrac}       
\usepackage{xcolor}         
\usepackage{hyperref}       
\hypersetup{colorlinks = true, allcolors = blue}
\usepackage[utf8]{inputenc} 
\usepackage[T1]{fontenc}    
\usepackage{url}            
\usepackage{booktabs}       
\usepackage{amsfonts,amsmath,mathrsfs}       
\usepackage{xcolor}         
\usepackage{multirow}

\usepackage{enumitem}
\setitemize[1]{itemsep=0pt,partopsep=0pt,parsep=\parskip,topsep=0pt}

\usepackage[accepted]{icml2023}

\usepackage{amsmath}
\usepackage{amssymb}
\usepackage{mathtools}
\usepackage{amsthm}

\usepackage[capitalize,noabbrev]{cleveref}

\theoremstyle{plain}
\newtheorem{theorem}{Theorem}[section]

\theoremstyle{definition}

\theoremstyle{remark}

\usepackage[textsize=tiny]{todonotes}

\icmltitlerunning{Sharpness-Aware Minimization Alone can Improve Adversarial Robustness}

\begin{document}

\twocolumn[
\icmltitle{Sharpness-Aware Minimization Alone can Improve Adversarial Robustness}



\icmlsetsymbol{equal}{*}

\begin{icmlauthorlist}
\icmlauthor{Zeming Wei}{equal,pku}
\icmlauthor{Jingyu Zhu}{equal,pku}
\icmlauthor{Yihao Zhang}{equal,pku}
\end{icmlauthorlist}

\icmlaffiliation{pku}{Peking University}

\icmlcorrespondingauthor{Zeming Wei}{weizeming@stu.pku.edu.cn}

\icmlkeywords{Machine Learning, ICML}

\vskip 0.3in
]



\printAffiliationsAndNotice{\icmlEqualContribution} 

\begin{abstract}
Sharpness-Aware Minimization (SAM) is an effective method for improving generalization ability by regularizing loss sharpness. In this paper, we explore SAM in the context of adversarial robustness. We find that using only SAM can achieve superior adversarial robustness without sacrificing clean accuracy compared to standard training, which is an unexpected benefit. We also discuss the relation between SAM and adversarial training (AT), a popular method for improving the adversarial robustness of DNNs. In particular, we show that SAM and AT differ in terms of perturbation strength, leading to different accuracy and robustness trade-offs. We provide theoretical evidence for these claims in a simplified model. Finally, while AT suffers from decreased clean accuracy and computational overhead, we suggest that SAM can be regarded as a lightweight substitute for AT under certain requirements.
Code is available at \url{https://github.com/weizeming/SAM_AT}.
\end{abstract}

\section{Introduction}
Sharpness-Aware Minimization (SAM)~\cite{foret2020sharpness} is a novel training framework that improves model generalization by simultaneously minimizing loss value and loss sharpness.
The objective of SAM is to minimize the \textit{sharpness} around the parameters, which can be formulated as 
\begin{equation}\label{eq1}
    \max_{\|\epsilon\|\le\rho} L(w+\epsilon)+\lambda \|w\|_2^2,
\end{equation}
where $L$ is the loss function, $w$ is the parameters of the model,
$\|w\|_2^2$ is the regularization term and $\rho$ controls the magnitude of weight perturbation. Intuitively, a larger $\rho$ leads to stronger weight perturbation and pushes the model to find a flatter loss landscape.
So far, SAM has become a powerful tool for enhancing the natural accuracy performance of machine learning models.

In this paper, we aim to explore SAM through the lens of adversarial robustness. Specifically, we study the robustness of SAM to defend against adversarial examples, which are natural examples with small perturbations that mislead the model into producing incorrect predictions~\cite{szegedy2013intriguing,goodfellow2014explaining}. The discovery of adversarial examples has raised serious concerns about the safety of critical domain applications~\cite{ma2019understanding}, and has attracted a lot of research attention in terms of defending against them. Currently, Adversarial Training (AT)~\cite{madry2017towards} has been demonstrated to be the most effective approach~\cite{athalye2018obfuscated} in improving the adversarial robustness of Deep Neural Networks (DNNs) among the various methods of defense. However, despite the success in improving adversarial robustness, there are still several defects remaining in adversarial training, such as decrease in natural accuracy~\cite{tsipras2018robustness}, computational overhead~\cite{shafahi2019adversarial}
, class-wise fairness~\cite{xu2021robust,wei2023cfa} and the absence of formal guarantees~\cite{wang2021betacrown, zhang2023using}.

Surprisingly, we find that models trained with SAM exhibit significantly higher adversarial robustness than those trained using standard methods, which is an unexpected benefit. Also, when comparing SAM to AT, SAM has the advantage of lower computational cost and no decrease in natural accuracy.
Based on the discussion above, we raise two research questions (\textbf{RQs}) in this paper:

\begin{itemize}
    \item \textbf{RQ1}: Why does SAM improve adversarial robustness compared to standard training?
    \item \textbf{RQ2}: Can SAM be used as a lightweight substitution for adversarial training?
\end{itemize}

To answer the two questions above, we first provide a comprehensive understanding of SAM in terms of adversarial robustness. Specifically, we present the intrinsic relation between SAM and AT that they both apply adversarial data augmentations to eliminate non-robust features~\cite{tsipras2018robustness}) from natural examples during the training phase.
As a result, both SAM and AT can effectively enhance the robustness of trained models, resulting in improved robust generalization ability. However, we also note that there are still several differences between SAM and AT. For instance, SAM adds adversarial perturbations \textit{implicitly}, while AT applies perturbations \textit{explicitly}. Additionally, the perturbation (attack) strength during training of SAM and AT differs, leading to different results in terms of natural and robust accuracy trade-offs.

Further, we verify the proposed empirical understanding with theoretical evidence in a simplified data model. Following the data distribution based on robust and non-robust features decomposition~\cite{tsipras2018robustness}, we show that both SAM and AT can improve the robustness of the trained models by biasing more weight on robust features. In addition, we also show that SAM requires a larger perturbation budget to achieve comparable robustness to AT,
which verifies our hypothesis that the perturbation strength of SAM is lower than AT.

Finally, we conduct experiments on benchmark datasets to verify our understanding. We find that models trained with SAM indeed outperform standard-trained models significantly in terms of adversarial robustness and also exhibit better natural accuracy. To sum up, our empirical and theoretical understanding can answer \textbf{RQ1}.

It is worth noting that, there still remains a large gap of robustness between SAM and AT.
However, the natural accuracy of AT is consistently lower than standard training, not to mention SAM. 
Meanwhile, SAM also outperforms AT in terms of computational cost.
Therefore, we finally answer \textbf{RQ2} with the conclusion that SAM can be considered a lightweight substitute for AT in improving adversarial robustness, under the following requirements:
(1) no loss of natural accuracy and
(2) no significant increase in computational cost.

To summarize, our main contributions in this paper are:
\begin{enumerate}
\item We point out that using SAM alone can notably enhance adversarial robustness without sacrificing clean accuracy compared to standard training, which is an unexpected benefit.

\item  We provide both empirical and theoretical explanations to clarify how SAM can enhance adversarial robustness. In particular, we discuss the relation between SAM and AT and demonstrate that they improve adversarial robustness by eliminating non-robust features.
However, they differ in perturbation strengths, which leads to different trade-offs between natural and robust accuracy.

\item We conducted experiments on benchmark datasets to verify our proposed insight. We also suggest that SAM can be considered a lightweight substitute for AT under certain requirements.

\end{enumerate}


\section{Background and related work}
\label{Section 2}
\subsection{Sharpness awareness minimization (SAM)}
In order to deal with the bad generalization problem in traditional machine learning algorithms, \cite{hochreiter1994simplifying,hochreiter1997flat} respectively attempt to search for flat minima and penalize sharpness in the loss landscape, which obtains good results in generalization~\cite{keskar2016large,neyshabur2017exploring,dziugaite2017computing}. 
Inspired by this, a series of works focus on using the concept of \textit{flatness} or \textit{sharpness} in loss landscape to ensure better generalization, \textit{e.g.} Entropy-SGD~\cite{chaudhari2019entropy} and Stochastic Weight Averaging (SWA)~\cite{,izmailov2018averaging}.
Sharpness-Aware minimization (SAM)~\cite{foret2020sharpness} also falls into this category, which simultaneously minimizes loss value and loss sharpness as described in (\ref{eq1}).

Theoretically, the good generalization ability of SAM is guaranteed by the fact that of the high probability, the following inequality holds:

\begin{equation}
    L_{\mathscr{D}}(\boldsymbol{w}) \leq \max _{\|\boldsymbol{\epsilon}\|_{2} \leq \rho} L_{\mathcal{S}}(\boldsymbol{w}+\boldsymbol{\epsilon})+h\left(\|\boldsymbol{w}\|_{2}^{2} / \rho^{2}\right),
\end{equation}

where set $ \mathcal{S} $ is generated from distribution $ \mathscr{D} $, $ h: \mathbb{R}_{+} \rightarrow \mathbb{R}_{+} $ is a strictly increasing function.


There are also many applications of SAM in other fields of research like language models \cite{bahri2021sharpness} and fluid dynamics \cite{jetly2022splash}, showing the scalability of SAM to various domains. In addition, many improvements of the algorithm SAM spring up, like Adaptive SAM (ASAM)~\cite{kwon2021asam}, Efficient SAM (ESAM) \cite{du2021efficient}, LookSAM~\cite{liu2022towards}, Sparse SAM (SSAM)~\cite{mi2022make}, Fisher SAM~\cite{kim2022fisher} and FSAM~\cite{zhong2022improving}, which add some modifications on SAM and further improve the generalization ability of the model. 

\subsection{Adversarial robustness}
The adversarial robustness and adversarial training has become popular research topic since the discovery of adversarial examples~\cite{szegedy2013intriguing,goodfellow2014explaining}, which uncovers that DNNs can be easily fooled to make wrong decisions by adversarial examples that are crafted by adding small perturbations to normal examples.
The malicious adversaries can conduct adversarial attacks~\cite{chen2023rethinking,wei2023weighted} by crafting adversarial examples, which cause serious safety concerns regarding the deployment of DNNs.
So far, numerous defense approaches have been proposed~\cite{papernot2016distillation,xie2019feature,bai2019hilbert,mo2022adversarial,chen2023robust}, among which
adversarial training (AT)~\cite{madry2017towards,wang2021convergence} has been considered as the most promising defending method against adversarial attacks, which can be formulated as
\begin{equation}
    \min_w \mathbb E_{(x,y)\sim \mathcal D}\max_{\|\delta\|\le\epsilon} L(w; x+\delta,y),
\end{equation}
where $\mathcal D$ is the data distribution, $\epsilon$ is the margin of perturbation, $w$ is the parameters of the model and $L$ is the loss function (\textit{e.g.} the cross-entropy loss).
For the inner maximization process, Projected Gradient Descent (PGD) attack is commonly used to generate the adversarial example:

\begin{equation}
x^{t+1}=\Pi_{\mathcal B(x,\epsilon)} (x^t+\alpha\cdot\text{sign}(\nabla_{x} \ell(\theta;x^t,y))),
\label{PGD}
\end{equation}
where $\Pi$ projects the adversarial example onto the perturbation bound $\mathcal B(x,\epsilon) = \{x':\|x'-x\|_p\le \epsilon\}$ and $\alpha$ represents the step size of gradient ascent.

Though improves adversarial robustness effectively, adversarial training has exposed several defects such as 
computational overhead~\cite{shafahi2019adversarial}, class-wise fairness~\cite{xu2021robust,wei2023cfa},
among which the decreased natural accuracy~\cite{tsipras2018robustness, wang2023simple} has become the major concern. 
It is proved that there exists an intrinsic trade-off between robustness and natural accuracy~\cite{tsipras2018robustness}, which can explain why AT reduces standard accuracy significantly.

In the context of adversarial robustness, there are also several works that attempt to introduce a flat loss landscape in adversarial training~\cite{wu2020adversarial,yu2022robust,yu2022robust2}. The most representative one is Adversarial Weight Perturbation (AWP)~\cite{wu2020adversarial}, which simultaneously adds perturbation on examples and feature space to apply sharpness-aware minimization on adversarial training.
However, AWP also suffers from a decrease in natural accuracy.
Also, the reason why a flat loss landscape can lead to better robustness has not been well explained.

To the best of our knowledge, we are the first to uncover the intrinsic relation between SAM and AT, and we reveal that SAM can improve adversarial robustness by implicitly biasing more weight on robust features.

\section{Empirical understanding}
\label{Section 3}

In this section, we introduce our proposed empirical understanding on the relation between SAM and AT, which can explain how SAM improves adversarial robustness.

Recall that the goal of SAM is to minimize the generalization error and loss sharpness simultaneously. 
 The sharpness term can be described as $\max\limits_{||\boldsymbol{\epsilon}||<\rho}[ L_S(\boldsymbol{w}+\boldsymbol{\epsilon})-L_S(\boldsymbol{w})],$ and the loss term is $L_S(\boldsymbol{w}).$ By combining the two terms, we get the objective of SAM is 
 \begin{equation}
 \min_{\boldsymbol w}\mathbb E_{(x,y)\sim \mathcal D}\max_{||\boldsymbol{\epsilon}||<\rho} L_S(\boldsymbol{w}+\boldsymbol{\epsilon};x,y).
 \end{equation}
 Also, recall that the objective of AT is
 \begin{equation}
     \min_{\boldsymbol w} \mathbb E_{(x,y)\sim \mathcal D}\max_{\|\delta\|\le\epsilon} L_s({\boldsymbol w}; x+\delta,y).
 \end{equation}

To illustrate their relation, we first emphasize that both techniques involve adding \textbf{perturbation} as a form of {data augmentation} for eliminating non-robust features~\cite{Ilyas2019AdversarialEA}. However, AT \textbf{explicitly} adds these perturbations to input examples, while SAM focuses on perturbing the parameters, which can be considered an \textbf{implicit} kind of data augmentation on the feature space. Therefore, both techniques involve perturbation on features, but in different spaces.

To be more specific and formal, we can derive our understanding with a middle linear layer in a model, which extracts feature $z$ from input $x$: $z = Wx$. In AT, we add perturbations directly to the input space, resulting in $x \gets x + \delta$. However, in SAM, the perturbation is not directly applied to the input space, but to the parameter space as $W \gets W + \delta$. This leads to $Wx + W\delta$ for input perturbation and $Wx + \delta x$ for parameter perturbation. Both perturbations can be seen as a form of data augmentation, with the former being more explicit and the latter being more implicit.

In addition, we discuss the attack (perturbation) strength of AT and SAM. For SAM, the perturbation is relatively more moderate, as its perturbations are injected in the feature space. However, this small perturbation is still helpful in improving robustness, since it can eliminate the non-robust features implicitly.
On the other hand, in order to achieve the best robustness by destroying the non-robust features, AT applies larger and more straightforward perturbations to the input space, leading to better robustness but a loss in natural accuracy.

Therefore, the difference and relation between SAM and AT can be considered as a trade-off between robustness and accuracy~\cite{zhang2019theoretically}. In summary, SAM applies small perturbations implicitly to the feature space to maintain good natural accuracy performance, while AT utilizes direct data augmentation magnitudes, which may result in a severe loss in natural accuracy. We provide more theoretical evidence for these claims in the next section.

\section{Theoretical analysis}
\label{Section 4}

In this section, we provide a theoretical analysis of SAM and the relation between SAM and AT.
Following the robust/non-robust feature decomposition~\cite{tsipras2018robustness},
we introduce a simple binary classification model, in which we show the implicit essential similarity and difference of SAM and AT.
We first present the data distribution and hypothesis space,  
then present how SAM and AT work in this model respectively, and finally discuss their relations.

\subsection{A binary classification model}
Consider a binary classification task that the input-label pair $(\boldsymbol x,y)$ is sampled from $\boldsymbol x\in\{-1,+1\}\times \mathbb R^{d}$ and $y\in\{-1,+1\}$, and the distribution $\mathcal{D}$ is defined as follows.
\begin{equation} 
\begin{aligned}
& y \stackrel{\text { u.a.r }}{\sim}\{-1,+1\}, x_{1}=\{\begin{array}{ll}
+y, & \text { w.p. } p, \\
-y, & \text { w.p. } 1-p,
\end{array} \\ &  x_{2}, \ldots, x_{d+1} \stackrel{i . i . d}{\sim} \mathcal{N}(\eta y, 1),
\end{aligned}
\end{equation}

where $p\in(0.5,1)$ is the accuracy of feature $x_1$, constant $\eta>0$ is a small positive number.
In this model, $x_1$ is called the \textit{robust feature}, since any small perturbation can not change its sign. 
However, the robust feature is not perfect since $p<1$. 
Correspondingly, the features $x_2,\cdots,x_{d+1}$ are useful for identifying $y$ due to the consistency of sign,
hence they can help classification in terms of natural accuracy.
However, they can be easily perturbed to the contrary side (change their sign) since $\eta$ is a small positive, which makes them called non-robust features~\cite{Ilyas2019AdversarialEA}.

Now consider a linear classifier model which predicts the label of a data point by computing $f_{\boldsymbol w}(\boldsymbol x) = \mathrm{sgn}(\boldsymbol w \cdot \boldsymbol x)$, and optimize the parameters $w_1,w_2,\cdots,w_n$ to maximize $\mathbb{E}_{x.y \sim \mathcal{D}} \mathbf{1}(f_{\boldsymbol w}(x) = y).$ 
In this model, given the equivalency of $x_i(i = 2,\cdots, n),$ we can set $w_2 = \cdots = w_n = 1$ by normalization without loss of generality. Therefore, the numerical value $w_1$ has a strong correlation with the robustness of the model.
Specifically, larger $w_1$ indicates that the model bias more weight on the robust feature $x_1$ and less weight on the non-robust features $x_2,\cdots, x_{d+1}$, leading to better robustness.

In the following, we discuss the trained model under standard training (ST), AT, and SAM respectively.
To make our description clear, we denote the loss function $\mathcal{L}(\boldsymbol x,y,w)$ as $1 - \mathrm{Pr}(f_w(\boldsymbol x) = y)$ and for a given $\epsilon>0$, we define the loss function of SAM $\mathcal{L}^{SAM} $ as $ \max_{|\delta| \le\epsilon}\mathcal{L}(\boldsymbol x,y,w+\delta). $

\subsection{Standard training (ST)}
We first show that there exists an optimal parameter $w_1^*$ under standard training in this model by the following theorem:
\begin{theorem}[Standard training]
    In the model above, under standard training, the optimal parameter value is 
    \begin{equation}
    w_1 ^{*}= \frac{\ln p - \ln (1-p)}{2\eta}.    
    \end{equation}
    \label{theorem ST}
\end{theorem}
\vspace{-21pt}
Therefore, $w_1^*$ can be regarded as the parameter $w_1$ returned by standard training with this model.

\begin{table*}[!t]
\caption{Natural and Robust Accuracy evaluation on \textbf{CIFAR-100} dataset.}
    \centering
    \begin{tabular}{c|c|cc|cc}  
    \toprule
       \multirow{2}{*}{Method}  & \multirow{2}{*}{Natural Accuracy} & \multicolumn{2}{c|}{$\ell_\infty$-Robust Accuracy} &\multicolumn{2}{c}{$\ell_2$-Robust Accuracy} \\
       &
       & $\epsilon=1/255$ &  $\epsilon=2/255$ &$\epsilon=16/255$ &  $\epsilon=32/255$\\
       \midrule
        ST &  76.9 & 13.6 & 1.7 & 44.5 & 21.2\\
        SAM ($\rho=0.1$) & 78.0 & 19.6 & 3.0 & 51.5 & 27.2\\
        SAM ($\rho=0.2$) & 78.5 & 23.1 & 4.2 & 54.2 & 31.3\\
        SAM ($\rho=0.4$) & \textbf{78.7} & \textbf{28.3} & \textbf{6.5} & \textbf{57.0} & \textbf{36.2}\\
        \midrule
        AT ($\ell_\infty$-$\epsilon=1/255$) & \textbf{73.1} & \textbf{60.4} & 46.6 & \textbf{67.4} & \textbf{61.5} \\
        AT ($\ell_\infty$-$\epsilon=2/255$) & 70.1 & 60.3 & 50.6 & 65.7 & 60.6 \\
        AT ($\ell_\infty$-$\epsilon=4/255$) & 66.2 & 59.3 & \textbf{52.0} & 62.8 & 58.9 \\
        AT ($\ell_\infty$-$\epsilon=8/255$) & 60.4 & 55.1 & {50.4} & 57.0 & 54.3 \\
        \midrule
        AT ($\ell_2$-$\epsilon=16/255$) & \textbf{74.8} & 52.8 & 31.4& 66.3 & 57.7 \\
        AT ($\ell_2$-$\epsilon=32/255$) & 73.2 & 57.4 & 40.6 & \textbf{67.1} & \textbf{61.0} \\
        AT ($\ell_2$-$\epsilon=64/255$) & 70.7 & 58.1 & 45.9 & 66.1 & 60.7 \\
        AT ($\ell_2$-$\epsilon=128/255$) & 67.4 & \textbf{58.2} & \textbf{48.7} & 63.9 & 60.4 \\
        \bottomrule
    \end{tabular}
    \label{tab:exp100}
\end{table*}

\subsection{Adversarial training (AT)}
Now let's consider when AT is applied. In this case, the loss function is no longer the standard one but the expected adversarial loss 
\begin{equation}
    \underset{(\boldsymbol x, y) \sim \mathcal{D}}{\mathbb{E}}\left[\max _{||\boldsymbol\delta||_{\infty} \le \epsilon} \mathcal{L}(\boldsymbol x+\boldsymbol\delta, y ; w)\right].
\end{equation}

Similar to standard training, there also exists an optimal parameter $w_1^{AT}$ returned by adversarial training,
which can be stated in the following theorem:

\begin{theorem}[Adversarial training]
In the classification problem above, under adversarial training with perturbation bound $\epsilon < \eta$, 
the adversarial optimal parameter value 
\begin{equation}
  w_1^{AT} = \frac{\ln p - \ln (1-p)}{2(\eta-\epsilon)}.  
\end{equation}
\label{theorem AT}
\end{theorem}

We can see that $w_1$ has been multiplied by $\frac{\eta}{\eta - \epsilon},$ which has increased the dependence on the robust feature $x_1$ of the classifier. 
This shows the adversarially trained model pays more attention to robustness compared to the standard-trained one, which improves adversarial robustness.

\subsection{Sharpness-Aware Minimization (SAM)}
Now we consider the situation of SAM. Recall that the optimizing objective of SAM is
\begin{equation}
    \underset{(x, y) \sim \mathcal{D}}{\mathbb{E}}\left[\max _{|\delta| \le \epsilon} \mathcal{L}(x, y ; w+\delta)\right].
\end{equation}
 We first explain why SAM could improve the adversarial robustness by proving that the parameter $w_1$ trained with SAM is also larger than $w_1^*$, which is stated as follows:

\begin{theorem}[Sharpness-aware minimization]
    In the classification problem above, the best parameter for SAM training $w_1^{SAM}$ satisfies that 
    \begin{equation}
        w_1^{SAM}> w_1^*.
    \end{equation}
    \label{SAM1}
\end{theorem}

From theorem \ref{theorem AT} and \ref{SAM1} we can see that both $w_1^{AT}$ and $w_1^{SAM}$ are greater than $w_1^{*}$, which indicates both SAM and AT improve robustness of the trained model.
However, the qualitative relation is not sufficient to quantify how much robustness SAM achieves compared to adversarial training, and we attempt to step further by quantitatively estimating the $w_1^{SAM}$ in the following theorem:

\begin{theorem}
    In the classification problem above, denote the best parameter for SAM training $w_1^{SAM}$. Suppose that $\epsilon$ is small, we have $w_1^{SAM}\approx w_1^{*}+\frac{2}{3}w_1^*\epsilon^{2}$.
    \label{SAM2}
\end{theorem}

\subsection{Relation between SAM and AT}

We further discuss the distinct attack (perturbation) strength between AT and SAM.
Recall that in our empirical understanding in  Section~\ref{Section 3}, the perturbation of SAM is more moderate than AT,
which can be interpreted as SAM focusing on natural accuracy more and robustness less in the robustness-accuracy trade-off.
Therefore, to reach the same robustness level (which is measured by the dependence on feature $x_1$, \textit{i.e.} the magnitude of $w_1$), SAM requires a much larger perturbation range, while for AT, less perturbation over $x$ is enough.
Theoretically, the following theorem verifies our statement:
\begin{theorem} 
\label{relation}
Denote the perturbation range $\epsilon$ of AT and SAM  as $\epsilon_{AT}$ and $\epsilon_{SAM}$, respectively. Then, when both methods return the same parameter $w_1$, we have the following relation between $\epsilon_{AT}$ and $\epsilon_{SAM}$:

\begin{equation}
2+\frac{3}{\epsilon_{SAM}^2}\approx\frac{2\eta}{\epsilon_{AT}}
\label{eq32}
\end{equation}
\end{theorem}

From theorem~\ref{relation}, we can identify the different perturbation strengths of AT and SAM.
It can be easily derive from Theorem \ref{relation} that $\epsilon_{SAM}$ is larger than $\epsilon_{AT}$ when (\ref{eq32}) holds, since we assume $\eta$ is a small positive, $\epsilon$ is small in theorem~\ref{SAM2} and $\epsilon_{AT}<\eta$ in theorem~\ref{theorem AT}.
Therefore, to gain the same weight $w_1$ on robust features $x_1$, $\epsilon_{AT}$ only need to be chosen much smaller than $\epsilon_{SAM}$. On the other hand, under the same perturbation bound  $\epsilon_{AT}=\epsilon_{SAM}$, the model trained under AT has larger parameter $w_1$ than SAM, hence it focuses on more robustness yet decreases more natural accuracy.

All proofs can be found in Appendix~\ref{proofs}.
To sum up, we can conclude that AT utilizes explicit and strong perturbations for denoising non-robust features, while SAM leverages implicit and moderate perturbations. This is consistent with our empirical understanding in Section~\ref{Section 3} and we also verify these claims with experiments in the following section.

\section{Experiment}
\label{Section 5}

\begin{table*}[!htbp]
\caption{Natural and Robust Accuracy evaluation on \textbf{ CIFAR-10} dataset.}
    \centering
    \begin{tabular}{c|c|cc|cc}  
    \toprule
       \multirow{2}{*}{Method}  & \multirow{2}{*}{Natural Accuracy} & \multicolumn{2}{c|}{$\ell_\infty$-Robust Accuracy} &\multicolumn{2}{c}{$\ell_2$-Robust Accuracy} \\
       &
       & $\epsilon=1/255$ &  $\epsilon=2/255$ &$\epsilon=16/255$ &  $\epsilon=32/255$\\
       \midrule
        ST &  94.6 & 39.6 & 8.9 & 76.1 & 51.7 \\
        SAM ($\rho=0.1$) & \textbf{95.6} & 45.1 & 9.4 & 81.0 & 56.3\\
        SAM ($\rho=0.2$) & 95.5 & 48.9 & 10.2 & 82.9 & 58.8\\
        SAM ($\rho=0.4$) & 94.7 & \textbf{56.1} & \textbf{15.6} & \textbf{84.0} & \textbf{64.4}\\
        \midrule
        AT ($\ell_\infty$-$\epsilon=1/255$) & \textbf{93.7} & 86.4 & 75.5 & \textbf{90.5} & 86.4  \\
        AT ($\ell_\infty$-$\epsilon=2/255$) & 92.8 & \textbf{87.4} & 79.6 & 90.3 & \textbf{86.9} \\
        AT ($\ell_\infty$-$\epsilon=4/255$) & 90.9 & 86.4 & \textbf{81.3} & 88.3 & 85.7\\
        AT ($\ell_\infty$-$\epsilon=8/255$) & 84.2 & 81.5 & 78.4 & 82.5 & 80.8\\
        \midrule
        AT ($\ell_2$-$\epsilon=16/255$) & \textbf{94.5} & 82.2 & 61.7 & 90.3 & 84.5 \\
        AT ($\ell_2$-$\epsilon=32/255$) & 93.7 & 84.9 & 70.9 & \textbf{91.0} & 86.7 \\
        AT ($\ell_2$-$\epsilon=64/255$) & 92.7 & 85.6 & 75.2 & 90.7 & \textbf{87.5} \\
        AT ($\ell_2$-$\epsilon=128/255$) & 90.2 & \textbf{85.7} & \textbf{78.4} & 89.6 & 87.1\\
        \bottomrule
    \end{tabular}
\vspace{-8pt}
    \label{tab:exp}
\end{table*}
In this section, we present our experimental results to verify our proposed understanding.

\subsection{Experiment set-up}
To demonstrate the effectiveness of SAM in improving adversarial robustness, we compare models trained with the standard SGD optimizer to those trained with SAM. We also discuss adversarial training. However, we consider the robustness obtained by AT as an upper bound rather than a baseline for SAM.

In our experiment, we train the PreActResNet-18 (PRN-18)~\cite{he2016identity} model on the CIFAR-10 and CIAR-100 datasets~\cite{krizhevsky2009learning} with Cross-Entropy loss for 100 epochs.
The learning rate is initialized as 0.1 and is divided by 10 at the 75th and 90th epochs, respectively.
For the optimizer, the weight decay is set to \texttt{5e-4}, and the momentum is set to 0.9.

For SAM, we select the perturbation hyper-parameter $\rho$ from the range $\{0.1,0.2,0.4\}$. And for AT,
we consider both $\ell_2$ and $\ell_\infty$ robustness and train 4 models with different perturbation bounds for the two kinds of norms, respectively.

As for robustness evaluation, we consider robustness under $\ell_\infty$-norm perturbation bounds $\epsilon\in\{1/255,2/255\}$ and
$\ell_2$-norm perturbation bounds $\epsilon\in\{16/255,32/255\}$.
The robustness is evaluated under a 10-step PGD attack.

For all models, we run the experiment three times independently and report the average result. We omit the standard deviations since they are small (less than 0.5\%) and do not affect our claims.

\subsection{Accuracy and robustness evaluation}
The results of the experiments conducted on the CIFAR-100 and CIFAR-10 datasets are presented in Table~\ref{tab:exp100} and Table~\ref{tab:exp}, respectively.

We first discuss the natural and robust accuracy performance of SAM.
From the tables, we can see that all the models trained with SAM exhibit significantly better natural accuracy and robustness compared to those trained with standard training (ST). In particular, higher robustness is achieved by using larger values of $\rho$ with SAM. For the CIFAR-100 dataset, the model trained with $\rho=0.4$ demonstrates even multiple robust accuracy than ST, and its natural accuracy is still higher than that of ST.
Compared to the improvement in natural accuracy (approximately 2\%), the increase of robustness is more significant (\textbf{more than 10\% in average}).
Similarly, for the CIFAR-10 dataset, the model trained with SAM also outperforms ST in terms of clean accuracy and exhibits significant higher robustness than ST.
Therefore, we can conclude that SAM with a relatively larger weight perturbation bound $\rho$ is a promising technique for enhancing the performance of models in terms of adversarial robust accuracy without sacrificing natural accuracy.

Regarding adversarially trained models, 
although there remains a large gap between the robustness obtained by SAM and AT, all adversarially trained models exhibit lower natural accuracy than standard training, not to mention SAM.
Particularly, for $\ell_\infty$-adversarial training, even training with perturbation bound $\epsilon=1/255$ decreases natural accuracy at 3.8\% for CIFAR-100 and 0.9\% for CIFAR-10 datasets, respectively.
And also note that the larger perturbation bound $\epsilon$ used in AT, the worse natural accuracy is obtained by the corresponding model. Therefore, a key benefit of using SAM instead of AT is that there is no decrease in clean accuracy.
Additionally, note that solving the PGD process in AT results in significant computational overhead. Specifically, since we use 10-step PGD, all AT experiments require 10 times more computational cost compared to ST, while SAM only requires 1 time more.

Based on the discussion above, we reach the conclusion that SAM-trained models perform significantly better robustness without decreasing any natural accuracy compared to standard training methods. 
Furthermore, another benefit of SAM is that it does not require significant computational resources. 
Therefore, we point out that SAM can serve as a lightweight alternative to AT, which can improve robustness without a decrease in natural accuracy and significant training overhead.

\section{Conclusion}
\label{Section 6}
In this paper, we show that using Sharpness-Aware Minimization (SAM) alone can improve adversarial robustness, and reveal the fundamental relation between SAM and Adversarial Training (AT). We empirically and theoretically demonstrate that both SAM and AT add perturbations to features to achieve better robust generalization ability. However, SAM adds moderate perturbations implicitly, while AT adds strong perturbations explicitly. Consequently, they lead to different accuracy and robustness trade-offs. We further conduct experiments on benchmark datasets to verify the validity of our proposed insight. Finally, we suggest that SAM can serve as a lightweight substitute for AT under certain requirements.


\bibliography{ref}
\bibliographystyle{icml2023}

\newpage
\appendix
\onecolumn

\section{Proofs}
\label{proofs}
\subsection{Proof for Theorem 1}
\textbf{Proof.} Due to symmetry, we only need to calculate the case of $y=1$ without loss of generality. From the distribution, we can easily derive that $x_2 + \cdots + x_{d+1} \sim \mathcal{N}(\eta d,d)$.

Thus\begin{equation}\begin{aligned}
   &\underset{\boldsymbol w}{\text{argmax}}\;\mathbb{E}_{\boldsymbol x.y \sim \mathcal{D}} \mathbf{1}(f_{\boldsymbol w}(\boldsymbol x) = y)\\
=& \underset{\boldsymbol w}{\text{argmax}}\; p\Pr(x_2 + \cdots + x_{d+1} > -w_1) + (1-p)\Pr(x_2 + \cdots + x_{d+1} > w_1)\\
=&  \underset{\boldsymbol w_1}{\text{argmax}}\;\frac{p}{\sqrt{2\pi d}}\int_{-w_1}^\infty e^{-(t-\eta d)^2/2d}dt +\frac{1-p}{\sqrt{2\pi d}}\int_{w_1}^\infty e^{-(t-\eta d)^2/2d}dt\\
:=& \underset{\boldsymbol w_1}{\text{argmax}}\;u(w_1).
\end{aligned}\label{eq7}\end{equation}

Then, the best parameter $w_1$ can be derived by $du/dw_1 = 0.$ The derivative is \begin{equation}\frac{du}{dw_1} = \frac{p}{\sqrt{2\pi d}} e^{-(w_1+\eta d)^2/2d} - \frac{1-p}{\sqrt{2\pi d}}e^{-(w_1-\eta d)^2/2d} = 0.\end{equation}
Solving this, we get the optimal value
\begin{equation}w_1^* = \frac{\ln p - \ln (1-p)}{2\eta}. \quad \square\end{equation}

\subsection{Proof for Theorem 2}

\textbf{Proof.} As $x_1$ has been chosen to be in $\pm 1$, the perturbation over $x_1$ has no influence on it, and we can just ignore it.
Therefore, to attack the classifier by a bias $\boldsymbol\delta$, to make the accuracy as small as possible, an intuitive idea is to set $\boldsymbol\delta$ which minimizes the expectation of $x_i(i = 2,\cdots,d+1),$ which made the standard accuracy smaller.
In fact, the expected accuracy is monotonically increasing about each $\delta_i, i = 2,\cdots,d+1.$ 
Thus, choosing $\delta = (0,-\epsilon , \cdots, -\epsilon )$ can be the best adversarial attack vector for any $w>0$.
In this situation, this equals $x_i'(i = 2,\cdots, d+1) \sim \mathcal{N}(\eta-\epsilon ,d).$ 
Therefore, similar to equation (\ref{eq7}), we can derive the train accuracy which is 
\begin{equation}
    v(w) = p\Phi((w+(\eta-\epsilon) d)/\sqrt{d}) + (1-p)\Phi((-w+(\eta-\epsilon) d)/\sqrt{d}).
\end{equation}

Here $\Phi(\cdot)$ is the cumulative distribution function of a standard normal distribution. Now we only need to solve equation $dv/dw = 0$.
Through simple computation, this derives that
\begin{equation}\begin{aligned}    
   & p\exp({-(w_1+(\eta-\epsilon)d)^2/2d})/\sqrt{2\pi d}
   = (1-p)\exp({-(w_1-(\eta-\epsilon)d)^2/2d})/\sqrt{2\pi d}.
\end{aligned}
\end{equation} 
Solving this equation, we finally get the optimal value $$w_1^{AT} = \frac{\ln p - \ln (1-p)}{2(\eta-\epsilon)}.\quad \quad \hfill \square$$

\subsection{Proof for Theorem 3}

\textbf{Proof.} 
Define the expected clean accuracy function 
\begin{equation}u(w) = \frac{p}{\sqrt{d}}\Phi((w+\eta d)/\sqrt{d}) + \frac{(1-p)}{\sqrt{d}}\Phi((-w+\eta d)/\sqrt{d}),\end{equation}
where $\Phi(\cdot)$ is the cumulative distribution function of a standard normal distribution and $w\in \mathbb{R}$. The derivative is
\begin{equation}\label{eq14}
    \begin{aligned}\frac{d u(w)}{dw}=&\frac{p}{\sqrt{2\pi d}}e^{-(w+\eta d)^{2}/2d}-\frac{1-p}{\sqrt{2\pi d}}e^{-(w-\eta d)^{2}/2d}\\
    =&\frac{1}{\sqrt{2\pi d}}e^{-(w^{2}+\eta^{2}d^{2})/d}(pe^{-w\eta}-(1-p)e^{w\eta})
    \end{aligned}
\end{equation}
From the theorem \ref{theorem ST} for standard training, we know $u(w)$ has only one global minimum $w_1^*$. Thus, from (\ref{eq14}) we know that $\frac{d u(w)}{w}<0$ if $w>w_1^*$ and $\frac{d u(w)}{w}>0$ if $w<w_1^*$.

In the SAM algorithm where we set $\lambda=0$ and with $\epsilon$ given, we know that 
\begin{equation}
    w_1^{SAM} = \underset{w}{\text{argmax}}\min_{\delta \in [-\epsilon,\epsilon] }u(w+\delta).
\end{equation}

It is easy for us to know that state that 
\begin{equation}
    \min_{\delta \in [-\epsilon,\epsilon] }u(w_1^{SAM}+\delta)=\min \{u(w_1^{SAM}-\epsilon) , u(w_1^{SAM}+\epsilon)\}.
\end{equation}

If $u(w_1^{SAM}-\epsilon) > u(w_1^{SAM}+\epsilon),$ then $w_1^{SAM}+\epsilon>w_1^*$. Since $\frac{d u(w)}{w}$ is continuous and locally bounded, there exists $\delta_{0}>0$ such that $u(w_1^{SAM}-\epsilon-\delta_0)>u(w_1^{SAM}+\epsilon)$ and $u(w_1^{SAM}+\epsilon-\delta_0)>u(w_1^{SAM}+\epsilon)$
 Thus, we have
\begin{equation}
\begin{aligned}
    &\min \{u(w_1^{SAM}-\epsilon-\delta_0) , u(w_1^{SAM}+\epsilon-\delta_0)\}>\min \{u(w_1^{SAM}-\epsilon) , u(w_1^{SAM}+\epsilon)\}.
    \end{aligned}
\end{equation}
Therefore
\begin{equation}
    \min_{\delta \in [-\epsilon,\epsilon] }u((w_1^{SAM}-\delta_0)+\delta)>\min_{\delta \in [-\epsilon,\epsilon] }u(w_1^{SAM}+\delta),
\end{equation}
which means that $w_1^{SAM}$ is not the optimal value we want.
Similarly we can disprove that $u(w_1^{SAM}-\epsilon) <u(w_1^{SAM}+\epsilon).$ Thus, $u(w_1^{SAM}-\epsilon)=u(w_1^{SAM}+\epsilon).$

From this, we know that
\begin{equation}
    \int_{w_1^{SAM}-\epsilon}^{w_1^*}\frac{du(w)}{dw}dw=-\int_{w_{1}^*}^{w_1^{SAM}+\epsilon}\frac{du(w)}{dw}dw.
\end{equation}
Using (\ref{eq14}), we have
\begin{equation}\label{eq20}
\begin{aligned}
    &\int_{w_1^{SAM}-\epsilon}^{w_1^*}e^{-w^{2}}(pe^{-w\eta}-(1-p)e^{w\eta})dw\\
    =&-\int_{w_{1}^*}^{w_1^{SAM}+\epsilon}e^{-w^{2}}(pe^{-w\eta}-(1-p)e^{w\eta})dw.
\end{aligned}
\end{equation}
If $w_1^{SAM}\leq w_1^*$, define $h=w_1^{SAM}+\epsilon-w_1^*$. Thus,$h\leq\epsilon\leq-w_1^{SAM}+\epsilon+w_1^*$. In (\ref{eq20}), we have
\begin{equation}
\begin{aligned}\label{eq21}
    &\int_{-h}^0 e^{-(w_1^*+w)^{2}}(pe^{-(w_1^*+w)\eta}-(1-p)e^{(w_1^*+w)\eta})dw\\
    \leq &\int_{w_1^{SAM}-\epsilon}^{w_1^*}e^{-w^{2}}(pe^{-w\eta}-(1-p)e^{w\eta})dw\\
    =&-\int_{w_{1}^*}^{w_1^{SAM}+\epsilon}e^{-w^{2}}(pe^{-w\eta}-(1-p)e^{w\eta})dw\\
    =&-\int_{0}^{h}e^{-(w_1^*+w)^{2}}(pe^{-(w_1^*+w)\eta}-(1-p)e^{(w_1^*+w)\eta})dw\\
    =&\int_{-h}^{0}e^{-(w_1^*-w)^{2}}(-pe^{-(w_1^*-w)\eta}+(1-p)e^{(w_1^*-w)\eta})dw.
\end{aligned}
\end{equation}
Since $w_1^*>0$, we can know that $e^{-(w_1^*+w)^{2}}>e^{-(w_1^*-w)^{2}}>0$ for $w\in[-h,0)$.

For the function $r(v):=(pe^{-v \eta}-(1-p)e^{v\eta})$ is monotonically decreasing and has one zero point $w_1^*$, thus $pe^{-(w_1^*+w)\eta}-(1-p)e^{(w_1^*+w)\eta}>0$ and $-pe^{-(w_1^*-w)\eta}+(1-p)e^{(w_1^*-w)\eta}>0$ for $w\in [-h,0]$. And
\begin{equation}
\begin{aligned}\label{eq22}
    &pe^{-(w_1^*+w)\eta}-(1-p)e^{(w_1^*+w)\eta}-(-pe^{-(w_1^*-w)\eta}+(1-p)e^{(w_1^*-w)\eta})\\
    =&(pe^{-w_1^* \eta}-(1-p)e^{w_1^*\eta})(e^{w\eta}+e^{-w\eta})\\
    =&0.
\end{aligned}
\end{equation}
Therefore, $\forall w\in [-h,0)$,
\begin{equation}
    \begin{aligned}
        & e^{-(w_1^*+w)^{2}}(pe^{-(w_1^*+w)\eta}-(1-p)e^{(w_1^*+w)\eta})
        > e^{-(w_1^*-w)^{2}}(-pe^{-(w_1^*-w)\eta}+(1-p)e^{(w_1^*-w)\eta}).
    \end{aligned}
\end{equation}
Combining this with (\ref{eq21}), we reach a contradiction. 

Thus, $w_1^{SAM}>w_1^*$. This ends the proof. $\hfill \square$

\subsection{Proof for Theorem 4}

\textbf{Proof.} 
We proceed our proof from (\ref{eq20}). Since we have proven that $w_1^{SAM}>w_1^*$, we suppose that $h=w_1^*-w_1^{SAM}+\epsilon<\epsilon$. Therefore, we get

\begin{equation}
    \begin{aligned}
        0=&\int_{-h}^0 e^{-(w_1^*+w)^2}(pe^{-(w_1^*+w)\eta}-(1-p)e^{(w_1^*+w)\eta})dw-\int_{-h}^0 e^{-(w_1^*-w)^2}(-pe^{-(w_1^*-w)\eta}+(1-p)e^{(w_1^*-w)\eta})dw\\
        &-\int_{h}^{2\epsilon-h} e^{-(w_1^*+w)^2}(-pe^{-(w_1^*+w)\eta}+(1-p)e^{(w_1^*+w)\eta})dw.
    \end{aligned}
\end{equation}

Since we only focus on $h$, we consider omitting the $o(h^3)$ terms in the calculation. To be more specific, $o(w^2)$ term in the integral symbol $\int_{-h}^0$ can be omitted and $o(w)$ or $o(h)$ term for $w$ and $h$ in the integral symbol $\int_{0}^{2\epsilon-2h}$ can also be omitted.\footnote{The validity of abandoning these high order terms can be seen from the result of the calculation, which shows that $\frac{h-\epsilon}{h}\to 0$.}

Combined with the proof in (\ref{eq22}) and the definition of $w_1^*$, and abandoning the high order terms, we can calculate the right-hand side as follows.
\begin{equation}\label{eq25}
    \begin{aligned}
        RHS=&\int_{-h}^0 (e^{-(w_1^*+w)^2}-e^{-(w_1^*-w)^2})\times
        (pe^{-(w_1^*+w)\eta}-(1-p)e^{(w_1^*+w)\eta})dw
        \\&+\int_{h}^{2\epsilon-h} e^{-(w_1^*+w)^2}(-pe^{-(w_1^*+w)\eta}+(1-p)e^{(w_1^*+w)\eta})dw\\\approx
        &\int_{-h}^0 e^{-(w_1^*)^2}(1-2w_1^* w-1-2w_1^* w+o(w))\times
        (pe^{-w_1^* \eta}(1-w\eta)-(1-p)e^{w_1^*\eta}(1+w\eta))dw\\&-
        \int_{0}^{2\epsilon-2h} e^{-(w_1^*)^2}(1-2w_1^*h)\times
        (-pe^{-w_1^*\eta}(1-h\eta)+(1-p)e^{w_1^*\eta}(1+h\eta))dw\\\approx
        &4e^{-(w_1^*)^2}w_1^*\int_{-h}^0 w^2\eta(-pe^{-w_1^* \eta}-(1-p)e^{w_1^*\eta})dw-e^{-(w_1^*)^2}\int_{0}^{2\epsilon-2h}\eta h(pe^{-w_1^*\eta}+(1-p)e^{w_1^*\eta})dw\\
        \approx& \frac{4}{3}e^{-(w_1^*)^2}w_1^*(pe^{-w_1^* \eta}+(1-p)e^{w_1^*\eta})h^3
        -2e^{-(w_1^*)^2}(\epsilon-h)\eta h(pe^{-w_1^*\eta}+(1-p)e^{w_1^*\eta})\\\approx
        & \frac{2}{3}e^{-(w_1^*)^2}(pe^{-w_1^*\eta}+(1-p)e^{w_1^*\eta})\eta(2w_1^*h^2-3(\epsilon-h)).
    \end{aligned}
\end{equation}
Since $RHS=0$, by solving the last equality in (\ref{eq25}) we get that
\begin{equation}
    \frac{\epsilon-h}{h}=\frac{2}{3}w_1^*h= o(1).
\end{equation}
Thus, the calculation and the abandoning of high-order terms in the calculation above are valid.
Since $h\to\epsilon$, we have
\begin{equation}
\begin{aligned}
    \epsilon-h=&\frac{2}{3}w_1^*h^2
    \approx\frac{2}{3}w_1^*\epsilon^{2}.
\end{aligned}
\end{equation}
Therefore, we finally draw conclusion that $w_1^{SAM}\approx w_1^{*}+\frac{2}{3}w_1^*\epsilon^{2}$.
$\hfill \square$

\subsection{Proof for Theorem 5}

\textbf{Proof.}
When both methods derives the same optimal value $w_1'$, denote the standard training optimal parameter $w_1^* = (\ln p -\ln (1-p))/2\eta$, we have \begin{equation}
    w_1^* (1+ \frac{2}{3}\epsilon_{SAM}^{2}) \approx w_1',
\end{equation}
and that
\begin{equation}
    \frac{\eta}{\eta-\epsilon_{AT}}w_1^*=w_1'.
\end{equation}
Thus, we have
\begin{equation}
    \frac{\eta}{\eta-\epsilon_{AT}}\approx1+ \frac{2}{3}\epsilon_{SAM}^{2}
\end{equation}
Solving this equation, we get relationship\begin{equation}
   \frac{2}{3}\epsilon_{SAM}^{2}\epsilon_{AT}+\epsilon_{AT}\approx\frac{2}{3}\eta\epsilon_{SAM}^{2}
\end{equation}
By dividing both sides with $\epsilon_{AT}\epsilon_{SAM}^2$, the relation in the theorem can be simply derived. This ends the proof. $\hfill \square$


\end{document}